\documentclass{isprs} % isprs class modified 23-04-2019 (Dennis Wittich)
\usepackage{subfigure}
\usepackage{setspace}
\usepackage{geometry} % added 27-02-2014 Markus Englich
\usepackage{epstopdf}
\usepackage[labelsep=period]{caption}  % added 14-04-2016 Markus Englich - Recommendation by Sebastian Brocks
\usepackage[british]{babel} 
\usepackage[hang]{footmisc}
\usepackage{hhline}
\usepackage{tabularray}
\usepackage{graphicx}
\usepackage{subcaption}
\usepackage{amsmath}
\usepackage{color}
\usepackage{tabularray}
\usepackage[hidelinks]{hyperref}

\usepackage{url} 

\usepackage[authoryear]{natbib}

\begin{document}

\title{Zero-shot detection of buildings in mobile LiDAR using Language Vision Model}
\date{}

% KAO: Remove extra spacing
\author{
 J. M. Goo\textsuperscript{1}, Z. Zeng\textsuperscript{1}, J. Boehm\textsuperscript{1}\thanks{Corresponding author}}

% KAO: Remove extra newline
\address{
	\textsuperscript{1 }Department of Civil, Environmental and Geomatic Engineering, University College London, Gower Street, London, WC1E 6BT UK\\
 \cr – \{june.goo.21, zichao.zeng.21, j.boehm\}@ucl.ac.uk\\
}

% KAO: Remove extra spacing
% Anonymous submissions, authors' names should not be visible

% To make it blind uncomment below \author and \address
% \author{***** (for review, names must be rendered anonymous)}

% KAO: Remove extra newline
% Anonymous submissions, authors' affiliations should not be visible
% \address{**** (for review, affiliations must be rendered anonymous)}

% If the corresponding author is NOT the final author, always add a % space before the subsequent comma, i.e.
% first author name\textsuperscript{a,}\thanks{Corresponding author} , % second author name \textsuperscript{b}, etc.
% thanks to Niclas Borlin 05-05-2016

\commission{XX, }{YY} %This field is optional. If filled, XX and YY should be replaced by adequate numbers. See https://www2.isprs.org/commissions/
\workinggroup{XX/YY} %This field is optional.
\icwg{}   %This field is optional.

% KAO: Use times symbol
\abstract{
Recent advances have demonstrated that Language Vision Models (LVMs) surpass the existing State-of-the-Art (SOTA) in two-dimensional (2D) computer vision tasks, motivating attempts to apply LVMs to three-dimensional (3D) data. While LVMs are efficient and effective in addressing various downstream 2D vision tasks without training, they face significant challenges when it comes to point clouds, a representative format for representing 3D data. It is more difficult to extract features from 3D data and there are challenges due to large data sizes and the cost of the collection and labelling, resulting in a notably limited availability of datasets. Moreover, constructing LVMs for point clouds is even more challenging due to the  requirements for large amounts of data and training time. To address these issues, our research aims to 1) apply the Grounded SAM through Spherical Projection to transfer 3D to 2D, and 2) experiment with synthetic data to evaluate its effectiveness in bridging the gap between synthetic and real-world data domains. Our approach exhibited high performance with an accuracy of 0.96, an IoU of 0.85, precision of 0.92, recall of 0.91, and an F1 score of 0.92, confirming its potential. However, challenges such as occlusion problems and pixel-level overlaps of multi-label points during spherical image generation remain to be addressed in future studies.
}


\maketitle
\section{Introduction}
\begin{figure*}[t]
  \centering
    \includegraphics[width=\linewidth]{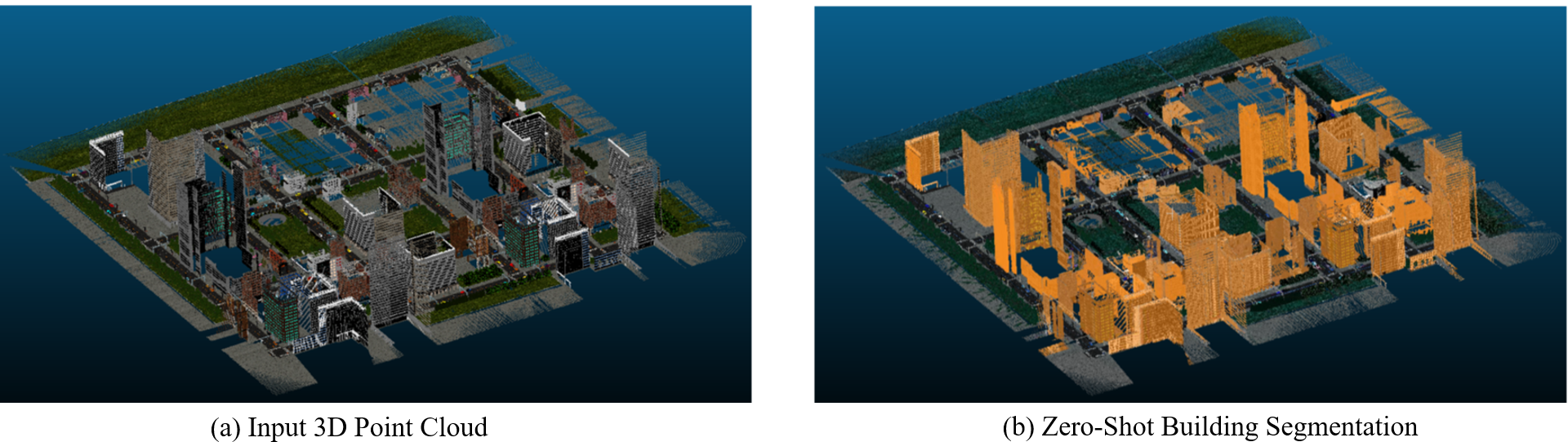}
    \caption{This figure illustrates the zero-shot building segmentation of the "SynthCity" dataset. (a) displays the original full point clouds used as input, while (b) shows the building detection results highlighted in yellow.
}
    \label{fig:full_synthcity_prediction}
\end{figure*}
With the advancement of 3D acquisition technologies such as LiDAR and other 3D sensors, along with reductions in price, the accessibility of point cloud data has become more affordable compared to the past. This has led to the emergence of multiple deep-learning techniques that can extract meaningful information from point clouds. \citep{qi2017pointnet, qi2017pointnet++, thomas2019kpconv, guo2021pct}. In the field of remote sensing and photogrammetry, building detection is one of the most fundamental problems for urban scenes as it forms the basis of 3D urban city model construction. Despite the remarkable progress in the field of deep learning, there are still significant obstacles to overcome when it comes to training deep learning models from scratch. One of the biggest challenges is the amount of data required for training, which can be particularly challenging in the case of point cloud datasets. Moreover, even when the dataset is available, training models on such vast amounts of data requires considerable computing power. In particular, this issue is a significant difficulty for the automation of 3D building detection, where data collection, processing, and time-consuming training are significant obstacles.

In order to address these limitations, several previous studies have proposed techniques such as Sim-to-Real domain transferable learning \citep{synthcity, xiao2022transfer, jin2022deformation, zhang2022learning}. This approach involves training models on synthetic datasets and effectively applying them to real-world datasets. Collecting and annotating real-world data is laborious and time-consuming, and training on synthetic dataset approach helps to avoid that. Moreover, synthesising data allows the generation of extensive data under various conditions such as different lighting and urban layouts. However, due to domain discrepancy, the models trained on synthetic datasets might struggle to generalise to real-world data due to differences in noise, density, and distribution of point clouds. Therefore, when training on synthetic datasets, caution must be taken to avoid overfitting. It is crucial to build models that possess robustness to accommodate the variations and imperfections in real-world data.

Alternatively, zero-shot transfer can be utilised for various computer vision tasks. Thanks to the advancement of Language Vision Models (LVMs), solving these tasks without the need for explicit training is now possible. LVMs, pre-trained on extensive datasets, consistently demonstrate high performance across different tasks. For instance, CLIP \citep{CLIP}, a prominent LVM, simultaneously trains an image encoder and a text encoder using an image-text paired dataset. Since the inception of the CLIP model, newer, more powerful models have emerged, such as OpenCLIP \citep{openclip}, ALIGN \citep{align}, and Flamingo \citep{flamingo}. Subsequently, LVM models have enhanced their robustness and adaptability through knowledge distillation for specific computer vision tasks like object detection \citep{grounding_dino, detpro, vldet} and semantic segmentation \citep{openseg, ov_seg}.

While Language Vision Models (LVMs) have demonstrated impressive performance in visual understanding without fine-tuning and with an open vocabulary, their reliance on extensive training data and computational resources remains a challenge, particularly in the context of 3D point clouds. To address these challenges, researchers have explored techniques such as leveraging pre-trained LVMs for point cloud understanding using multi-view approaches \citep{openmask3d, openscene}, knowledge distillation \citep{pointCLIP, pointclip2}, or projections \citep{pointCLIP} instead of training LVMs directly on point cloud data from scratch.

In this study, we aim to achieve two main goals. Firstly, we seek to perform 3D point cloud segmentation by adapting an LVM initially designed for 2D computer vision. This adaptation is facilitated through a 3D-to-2D projection method, eliminating the need for pre-training or fine-tuning. Secondly, we analyse the usefulness of this approach by conducting experiments on synthetic data to assess its ability to bridge the domain gap between synthetic and real-world datasets.

\section{Backgrounds and Previous Works}
\subsection{Language Vision Models for Point Clouds}
Thanks to the high impact of LVMs on the field of computer vision, many research efforts have been made to incorporate LVMs into 3D data processing. However, due to the significant amount of training time and data required to construct LVMs, most progress has been made only in 2D data. These drawbacks are more emphasised when it comes to 3D data, which are much larger in size and require longer training times. Therefore, many works have started investigating ways to utilise existing 2D LVMs for 3D data understanding rather than building LVMs specifically for 3D data, such as point clouds, meshes and voxels.

One method to enable 3D recognition with CLIP-based models is by projecting point clouds onto 2D images. PointCLIP \citep{pointCLIP} enables cross-modality zero-shot recognition on point clouds without prior 3D training by leveraging multi-view simple projection to transfer pre-trained 2D knowledge from CLIP to the 3D domain. Under a lightweight inter-view adapter under few-shot settings, PointCLIP enhances classification performance. However, it still exhibits low performance in zero-shot transfer scenarios. Nevertheless, PointCLIP-V2 \citep{pointclip2} addresses these shortcomings by replacing the multi-view simple projection with a realistic projection and employing prompt engineering with GPT-3 \citep{gpt3}, enhancing performance not limited to classification but also part segmentation and object detection. 

An alternative method involves aligning features from point cloud encoders with CLIP representations. For instance, ULIP \citep{xue2023ulip} introduces a method that trains triplets consisting point clouds, images, and texts using a limited set of synthesised triplets to align with CLIP image-text space. \cite{liu2024openshape} enhance 3D representations by aggregating multiple 3D datasets and refine noisy text descriptions through the utilisation of a powerful large language model, GPT-4 \citep{openai2023gpt4}.

Success in point cloud classification leads to more complex tasks such as object detection and segmentation. OV-3DET \citep{OV_3DET} introduces a novel de-biased triplet cross-modal contrastive learning to connect image, point-cloud, and text modalities for improved performance with LVMs. OpenMask3D \citep{openmask3d} and OpenScene \citep{openscene} models enable Open-Vocabulary 3D scene understanding utilising the pre-trained Image-Text Embedding model, CLIP. These models, based on multi-view scene understanding, are more complex and error-prone as they require precise camera calibration in data integration. Moreover, they are limited to indoor scenes, raising uncertainty about their robustness for outdoor scenes.
\subsection{Building Segmentation from Point Clouds}
Building segmentation tasks in recent studies are often separated based on the method used to collect point clouds. Aerial point clouds are used for building detection and segmentation, while point clouds collected from terrestrial LiDAR or Mobile Laser Scanning (MLS) are more likely to be used for building part segmentation of the facade. Building detection from point clouds is an area that has received limited research attention. Even when studied, most models are trained on aerial point clouds rather than using MLS.

Several previous studies have used classical geometric approaches or non-deep learning methods for building detection. For example, \cite{RANSAC_DBSCAN_BD} proposed a method that uses RANSAC and DBSCAN algorithms to detect building footprints. Additionally, \cite{density_BD} introduced a new approach that involves removing trees and vegetation and analysing the density within a cube in both 2D and 3D dimensions to identify buildings from the LiDAR point cloud.

Nowadays, deep learning models have enabled the extraction of features and solving specific tasks in point cloud understanding. \cite{UNet_LIDAR_PHOTO} proposed a methodology for building extraction through the fusion of LiDAR and photogrammetric point clouds, applying U-Net deep learning model segmentation.

A recent study, published by \cite{DGCNN_pointnet_aerial}, utilised PointNet \citep{qi2017pointnet} and Dynamic Graph Convolutional Neural Network (DGCNN) \citep{dgcnn} to detect buildings in cities of Indonesia. They collected the necessary LiDAR data using unmanned aerial vehicles, resulting in a promising approach to building detection. The Damage-Sensitive Network (DS-Net)\citep{DC-Net} is introduced as a specialised method for identifying collapsed buildings with the Laplacian Unit (LU).

\section{Methodology}
\label{methodology}
In our work we use 2D rendered views of the 3D point clouds to perform scene understanding (as reviewed in \cite{rs11121499}). This projection and back-projection process follows work by \cite{UCL_panoramic_projection,panoramic_images_poux_journal,panoramic_images_poux}. Building on the 2D projection, we approach our task of building segmentation using the combination of two 2D foundation models: Segment-Anything Model (SAM) \citep{SAM} and Grounding DINO \citep{grounding_dino}. Combining these two models was introduced by IDEA research as Grounded Segment-Anything (or Grounded SAM)\citep{Grounded-SAM_Contributors_Grounded-Segment-Anything_2023}.

\begin{figure}[t]
  \centering
    \includegraphics[width=1\linewidth]{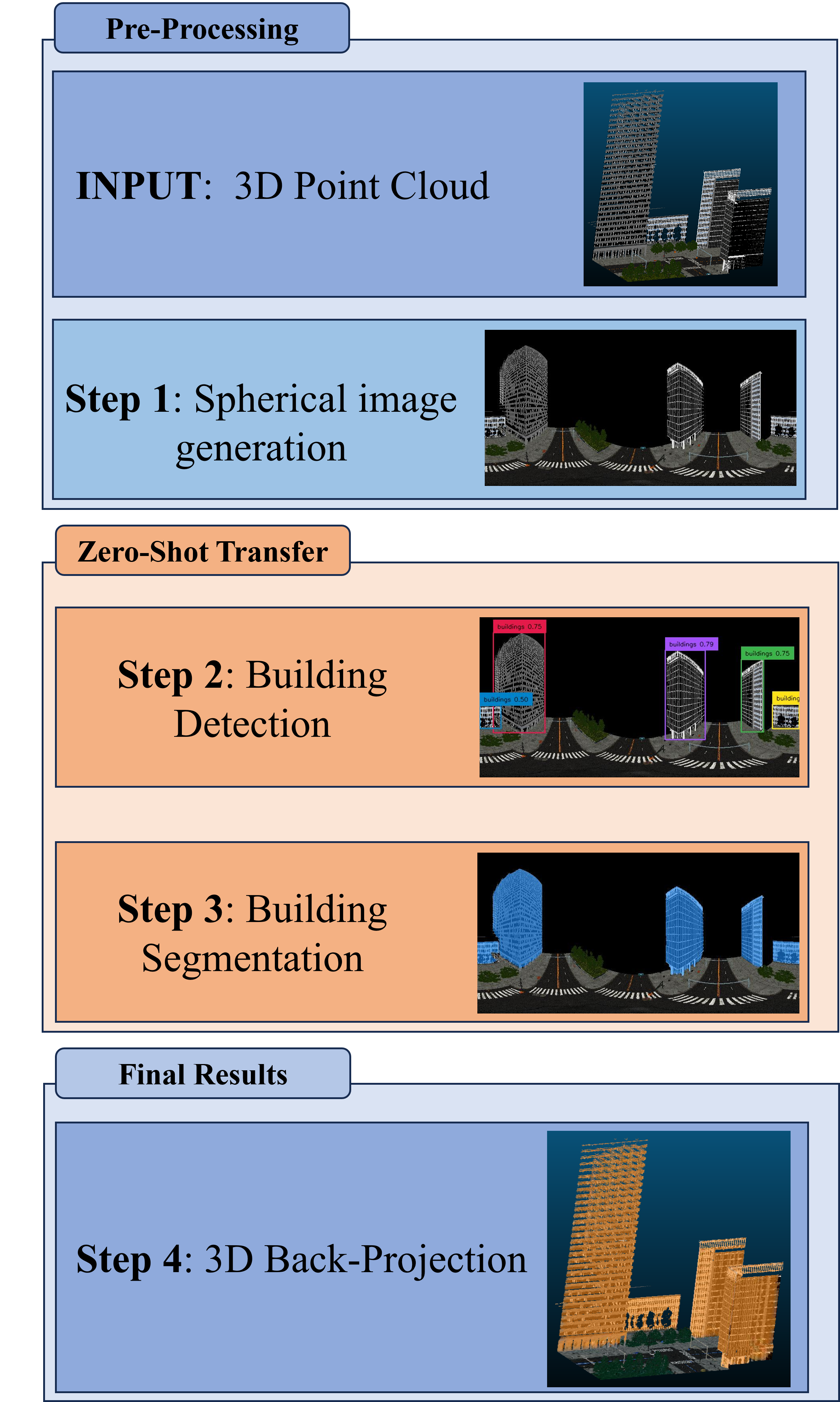}
    \caption{The diagram illustrates the steps of the methodological workflow. The figures on the right side are the examples of each step.}
    \label{fig:workflow}
\end{figure}

Our indirect building segmentation method consists of 4 steps as shown in Figure \ref{fig:workflow}. Each step will be introduced in the following sections.

\subsection{Spherical Image Generation}
In order to generate a spherical image from the point clouds, we first normalise the point $P(x, y, z)$ by the reference point $O(x_0, y_0, z_0$) as the following equation: 
\begin{align}
    x^{'} = x - x_0,\quad y^{'} = y - y_0,\quad z^{'} = z - z_0
\label{eq:normalization}
\end{align}
We set the reference coordinate in the middle of the road or at an intersection.

\begin{figure}[t]
  \centering
    \includegraphics[width=0.5\linewidth]{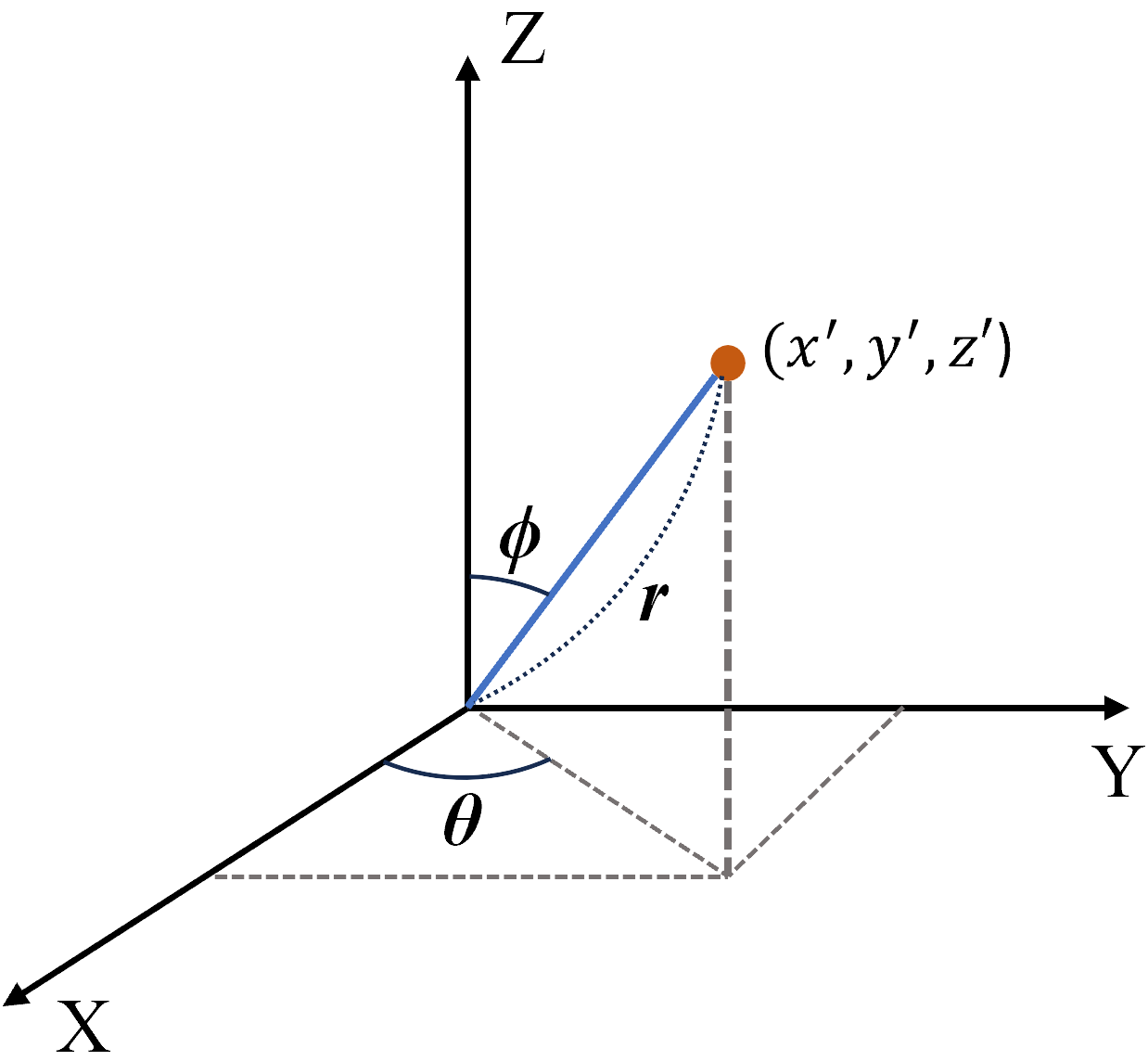}
    \caption{Principles of Spherical Coordinate Generation}
    \label{fig:spherical_coordinates}
\end{figure}

In the next step, we compute the spherical coordinates ($\theta$, $\phi$) for all normalised point clouds. The Figure \ref{fig:spherical_coordinates} shows the visual explanation of spherical coordinates, where $r$ is the radius from the origin to the point, $\theta$ is the azimuthal angle, and $\phi$ is the polar angle. Each coordinate is derived by the following equations:
\begin{align}
    r = \sqrt(x'^2 + y'^2 + z'^2) \\
    \theta = arctan2(y', x') \\
    \phi = arccos(z', r)
\label{eq:spherical_coordinates}
\end{align}
where $x', y', z'$ are the normalised coordinates from equation \ref{eq:normalization}

Lastly, we project these spherical coordinates to the 2D plane (image), which is also called an Equirectangular Projection. Each x and y coordinate in the image is determined by the following equations:
\begin{align}
    x = \frac{\theta+\pi}{2\pi} \times W \\
    y = \frac{\phi}{\pi} \times H
\label{eq:equirectangular_projection}
\end{align}
where $W$ and $H$ represent the width and height of the image to project respectively.

Step 1 in Figure \ref{fig:workflow} shows the result of the spherical image generation procedure of the input 3D point clouds. This figure represents a spherical image generated from a reference point located at the centre of an intersection on the road.

\subsection{Building Bounding-Box Detection}
For building detection, we use an open-set object detection model, Grounding DINO. Grounding DINO is a model that merges the image feature encoder DINO with grounded pre-training to detect a wide range of objects using language inputs \citep{grounding_dino}. This model integrates language for open-set detection, dividing the detection process into feature enhancer, language-guided query selection, and a cross-modality decoder \citep{grounding_dino}. We use the model for our task, by using the text prompt "buildings" to obtain bounding boxes of the buildings in the generated image. Step 2 of Figure \ref{fig:workflow} shows an example of the results of building detection using Grounding DINO.

\subsection{Building Segmentation}
We proceed to use SAM for the segmentation step. SAM is a foundation model for image segmentation, using input prompts such as points or masks \citep{SAM}. It also explores zero-shot segmentation from free-form text \citep{SAM}. However, we specifically focus on using the segmentation capabilities of SAM with bounding box inputs. For each bounding box detected in the previous step, we run SAM to segment all the pixels that belong to the building category. In Step 3 of Figure \ref{fig:workflow}, blue-coloured areas represent the segmented pixels corresponding to the buildings.

\subsection{Back-Projection}
After segmenting the building image, we perform a back-projection of the segmented image onto the 3D point cloud. To achieve this, we save the mapping details during the generation of the spherical image. Later, we load these mapping details and replace the RGB data of the initial point clouds with the points that correspond to the segmented pixels.

\section{Experiments}
We use an NVIDIA GeForce RTX 3070 with 8 GB for running Grounding DINO and SAM.

\subsection{Performance Metrics}
In order to measure model performance, we evaluate a range of commonly used metrics including Accuracy, Precision, Recall, F-1 Score, and Intersection over Union (IoU). IoU is typically used specifically for segmentation tasks. For detailed formulas refer to e.g. \cite{goodfellow2016deep, loss_metrics}

\subsection{Synthcity Dataset}
SynthCity \citep{synthcity} is an open-source dataset that represents an entire city in the form of a large-scale synthetic point cloud. The dataset includes 9 sub-areas and 9 label categories. SynthCity is specifically designed for pre-training deep learning models, enabling generalisation and expansion of their usage to real-world data. The dataset comprises 367 million completely labelled points with RGB features, making it a valuable resource for researchers and practitioners in this field. All points are labelled with one of 9 categories: Building, Car, Natural Ground, Ground, Pole Like, Street Furniture, Tree, and Pavement. 

To avoid reaching computational resource limits, each sub-area in SynthCity is further divided into 4 sub-sub areas. Our dataset excludes 2 sub-sub areas of Area 7 and 1 sub-sub area of Area 9 as they contain neither roads nor buildings. As mentioned in Section \ref{methodology} earlier, we arbitrarily selected centre points to generate spherical images in each sub-sub area. When making these selections, we imposed a condition that the centre point must first align with the roads. Additionally, if the area includes intersections, we designated the centre point to coincide with the centre of the intersection.

\section{Experimental Results}
We present the comprehensive calculations of the error statistics for all sub-sub areas from Area 1 to 9.
\subsection{Quantitative Evaluation}
\begin{figure*}[t]
  \centering
    \includegraphics[width=0.8\linewidth]{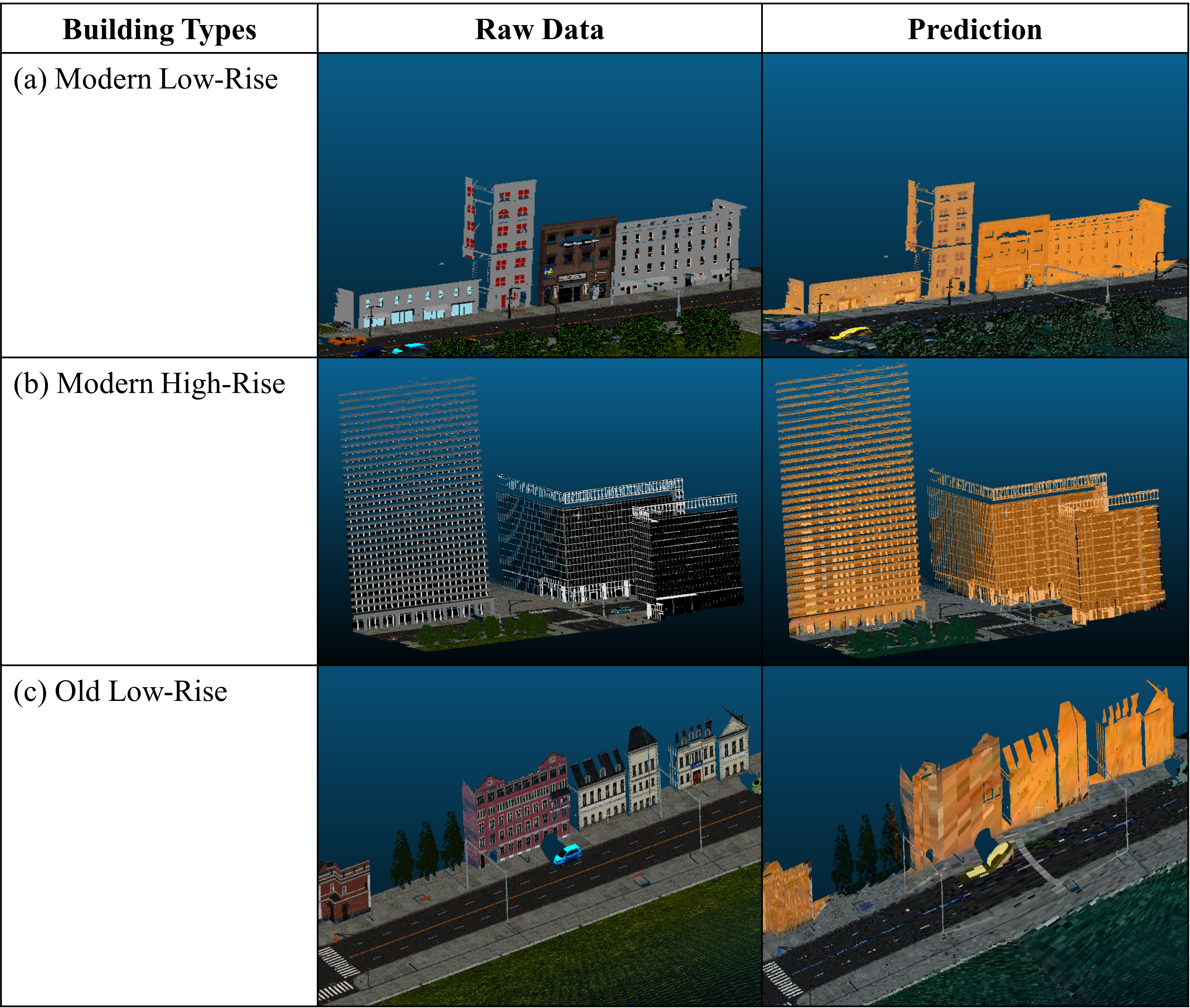}
    \caption{The figure shows the results of the segmentation of buildings using zero-shot prediction across different types of buildings. The results indicate that this method is highly effective in achieving accurate building segmentation regardless of building type}
    \label{fig:results_comparison}
\end{figure*}

For the proper assessment of our approach, we compute five metrics. Table \ref{table:quantitative_results} displays an overall Accuracy of 0.96, IoU of 0.85, Precision of 0.92, Recall of 0.91, and F1-Score of 0.92 across all areas. Since our method has not been benchmarked against other approaches, we cannot claim it to be the superior choice. Nevertheless, given its highly encouraging outcomes, we wish to emphasise the versatility and resilience of LVMs when applied to 3D synthetic datasets using spherical images. 

\begin{table}[t]
\centering
\begin{tblr}{
  cells = {c},
  vline{2} = {-}{},
  hline{2,11} = {-}{},
}
\textbf{Area}   & \textbf{Accuracy} & \textbf{IoU}  & \textbf{Precision} & \textbf{Recall} & \textbf{F-1} \\
\textbf{Area 1} & 0.86              & 0.67          & 0.72 & 0.92            & 0.8                \\
\textbf{Area 2} & 0.98              & 0.90          & 0.97 & 0.93            & 0.95               \\
\textbf{Area 3} & 0.97              & 0.91          & 0.98 & 0.93            & 0.96               \\
\textbf{Area 4} & 0.93              & 0.79          & 0.98 & 0.8             & 0.88               \\
\textbf{Area 5} & 0.97              & 0.88          & 0.95 & 0.92            & 0.94               \\
\textbf{Area 6} & 0.98              & 0.94          & 0.98 & 0.96            & 0.97               \\
\textbf{Area 7} & 0.97              & 0.76          & 0.97 & 0.78            & 0.86               \\
\textbf{Area 8} & 0.98              & 0.86          & 0.96 & 0.89            & 0.92               \\
\textbf{Area 9} & 0.99              & 0.96          & 1.00 & 0.96            & 0.98               \\
\textbf{Total}  & \textbf{0.96}     & \textbf{0.85} & \textbf{0.92 }& \textbf{0.91}   & \textbf{0.92}      
\end{tblr}
\caption{Quantitative Analysis of 9 Sub-Areas}
\label{table:quantitative_results}
\end{table}

\subsection{Qualitative Evaluation}
The left image in Figure \ref{fig:full_synthcity_prediction} illustrates the complete raw SynthCity dataset, while the right image represents the entire building prediction of the dataset. Our method shows remarkable performance not only in terms of quantitative analysis but also in qualitative analysis, as seen in the full dataset view.

The Figure \ref{fig:results_comparison} displays a detailed analysis of the different types of buildings. (a) and (b) highlight the detection of high and low-rise modern buildings, respectively. The results obtained demonstrate the high accuracy of our method in detecting buildings with different heights. Furthermore, (c) illustrates that our method can detect not only modern-style buildings but also old European-style ones, indicating the robustness and versatility of our approach.

\subsection{False Positive Analysis}
\begin{table}[t]
\centering
\resizebox{\linewidth}{!}{%
\begin{tblr}{
  cells = {c},
  cell{2}{5} = {fg=red},
  cell{3}{4} = {fg=red},
  cell{4}{4} = {fg=red},
  cell{5}{4} = {fg=red},
  cell{6}{7} = {fg=red},
  cell{7}{8} = {fg=red},
  cell{8}{7} = {fg=red},
  cell{9}{4} = {fg=red},
  cell{10}{4} = {fg=red},
  cell{11}{5} = {fg=red},
  vline{2} = {-}{},
  hline{2,11} = {-}{},
}
                & \textbf{Car} & {\textbf{Natural}\\\textbf{Ground}} & \textbf{Ground} & \textbf{Road} & {\textbf{Street}\\\textbf{Furniture}} & \textbf{Tree} & \textbf{Pavement} \\
\textbf{Area 1} & 0.024        & 0.011                               & 0.012           & 0.846         & 0.016                                 & 0.009         & 0.076             \\
\textbf{Area 2} & 0            & 0.108                               & 0.586           & 0             & 0.015                                 & 0.277         & 0.01              \\
\textbf{Area 3} & 0            & 0.146                               & 0.509           & 0             & 0.033                                 & 0.298         & 0.014             \\
\textbf{Area 4} & 0            & 0.037                               & 0.577           & 0             & 0.051                                 & 0.065         & 0.268             \\
\textbf{Area 5} & 0            & 0.133                               & 0.287           & 0             & 0.011                                 & 0.397         & 0.192             \\
\textbf{Area 6} & 0            & 0                                   & 0.362           & 0.012         & 0.049                                 & 0.126         & 0.441             \\
\textbf{Area 7} & 0            & 0                                   & 0.445           & 0             & 0                                     & 0.528         & 0.027             \\
\textbf{Area 8} & 0            & 0.032                               & 0.853           & 0             & 0.003                                 & 0.098         & 0.014             \\
\textbf{Area 9} & 0            & 0                                   & 1               & 0             & 0                                     & 0             & 0                 \\
\textbf{Total}  & 0.018        & 0.031                               & 0.137           & 0.617         & 0.019                                 & 0.076         & 0.098             
\end{tblr}
}
\caption{False Positive ratio for each category}
\label{table:FPR}
\end{table}

\begin{figure}[t]
  \centering
    \includegraphics[width=\linewidth]{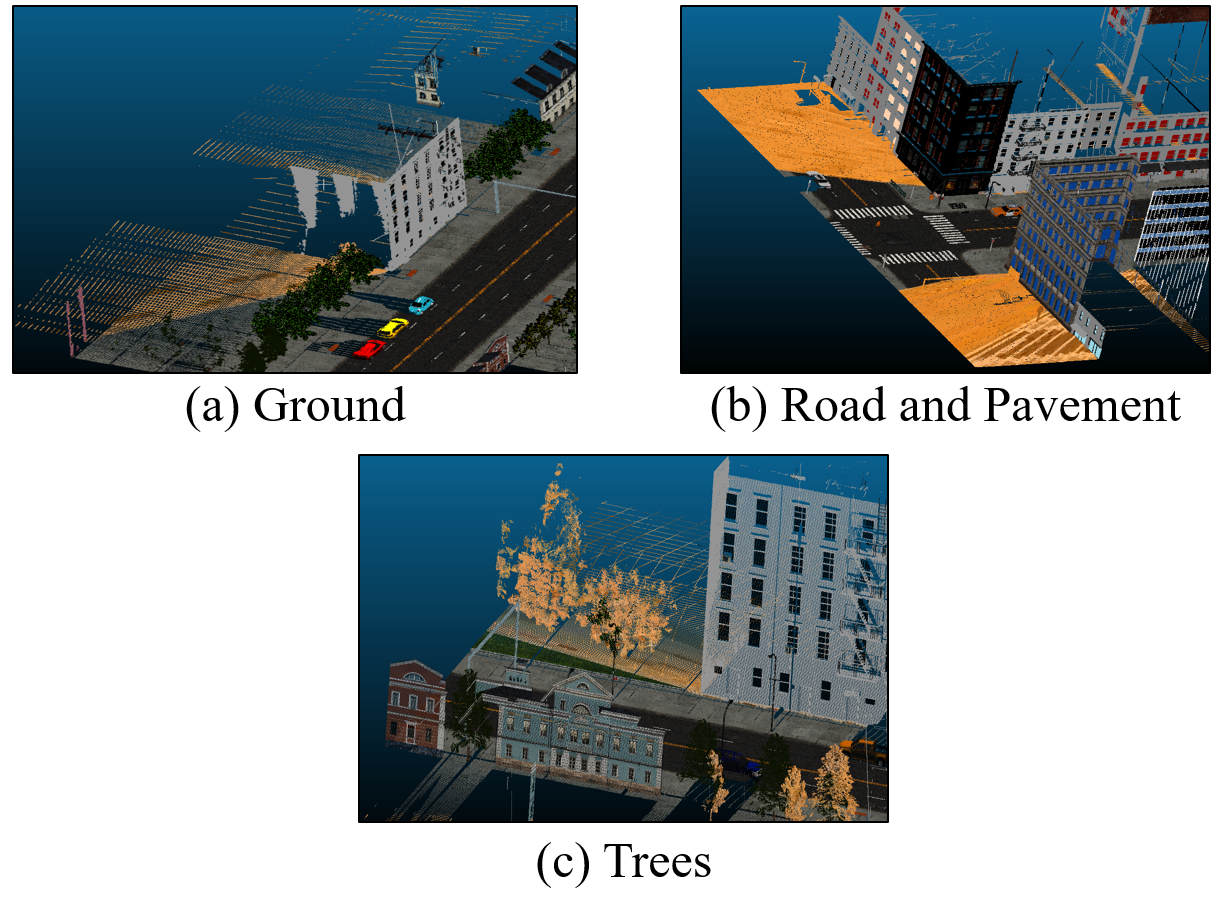}
    \caption{The figure shows in orange the false positive building points for different object categories. For clarity true positives are not highlighted.}
    \label{fig:FP_visualization}
\end{figure}

\begin{figure}[t]
  \centering
    \includegraphics[width=\linewidth]{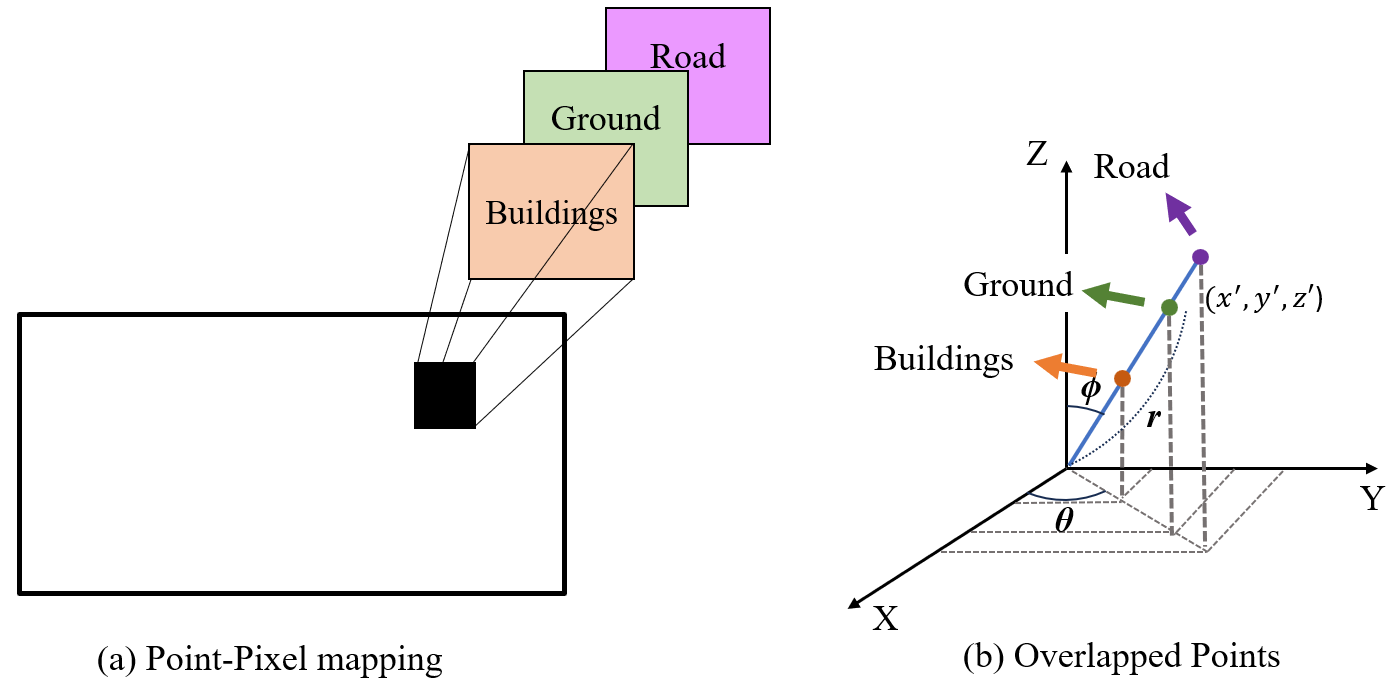}
    \caption{The figure displays the process of mapping that leads to the occurrence of false positives.}
    \label{fig:mapping_visualization}
\end{figure}

We have conducted a False Positive analysis of our method and identified a pattern of misclassified points. Analysis from Table \ref{table:FPR} revealed that flat surfaces such as roads, grounds, and pavements often get incorrectly classified as buildings because some points classified as buildings pass through during the back- projection process. This misclassification is not limited to flat surfaces. For instance, tall and large-scale objects, such as trees, can also be labelled as buildings. Because trees typically stand close together, some points predicted as buildings might be projected close to points belonging to the tree category.

From the perspective of visualization, as shown in the Figure \ref{fig:FP_visualization}, false positives occur when we back-project from the mapping of the spherical image. In (a), the segmented area is incorrectly back-projected to the ground behind the buildings, while in (b) and (c), it is incorrectly back-projected to roads, pavements, and trees respectively. They all have a common phenomenon: all false positives are generated behind the buildings. These results are generated because, for each pixel of the spherical image, our method saves more than one point, each with different labels. As in Figure \ref{fig:mapping_visualization} (a), while we generate Point-Pixel mapping, multiple points with labels could be saved in one pixel. This happens due to several reasons: First, when we generate spherical coordinates, the points with the same $\theta$ (azimuthal angle) and $\phi$ (polar angle) but different $r$ (distances from the origin) will be mapped to the same pixel. This is because the equirectangular projection only considers the angular components of the spherical coordinates and not the radial distance as in the Figure \ref{fig:mapping_visualization} (b). Secondly, the SAM model segments buildings, however, some contour pixels are mistakenly classified as building labels. The pixels that were incorrectly segmented will be projected as buildings, including any points that were within those pixels.

\section{Conclusion}
In our study, we experimented with a training-free zero-shot transfer approach by adopting spherical image generation using the 2D-3D projection method for building segmentation tasks, applying a LVM. We integrated the powerful Open-vocabulary segmentation model, Grounded SAM, which combines the Open-vocabulary object detection model called Grounding DINO with the segmentation-enhanced Segment-Anything Model (SAM). Instead of the commonly used Multi-View projection, we utilised a 360-degree spherical projection, which alleviates the complexities of considering various orientations associated with multi-view projection, offering a relatively simple and intuitive approach. Despite the shape distortion challenges of spherical projection, Grounded SAM overcame building shape distortions and confirmed the effectiveness of LVM in accurately detecting buildings. Qualitative analysis revealed the successful detection of buildings regardless of their construction era or height. Quantitatively, the model exhibited excellent performance across all metrics, showing the robustness and versatility of LVM. However, a limitation of spherical image generation is the potential for false positives behind buildings due to the ability to store multiple labels for points in a single pixel. Nevertheless, despite these limitations, the use of LVM for zero-shot transfer allowed computational freedom and demonstrated effective operation even on synthesised datasets, confirming its domain transferability. These promising results could potentially address the lack of point cloud training datasets and compute-intensive training.

\section{Acknowledgment}
J. M. Goo and Z. Zeng are supported by the Engineering and Physical Sciences Research Council through an industrial CASE studentship with Ordnance Survey (Grant number EP/X524840/1).

{
	\begin{spacing}{1.17}
		\normalsize
		\bibliography{ISPRS_2024_Goo} % Include your own bibliography (*.bib), style is given in isprs.cls
	\end{spacing}
}

\end{document}